\documentclass[letterpaper, 10 pt, conference]{ieeeconf}  

\IEEEoverridecommandlockouts                              

\usepackage{pdfpages}

\title{\LARGE \bf
IntentionVLA: Generalizable and Efficient Embodied Intention Reasoning for Human-Robot Interaction 
}

\author{
Yandu Chen$^{1}$, Kefan Gu$^{2}$, Yuqing Wen$^{3}$, Yucheng Zhao$^{4}$\textsuperscript{\textdagger}, Tiancai Wang$^{4}$, Liqiang Nie$^{1\ddag}$
\thanks{ \ \ This work was done during the internship at Dexmal.}
\thanks{\textdagger \ Project lead.}
\thanks{\ddag \ Corresponding author.}
\\[0.5em] 
$^{1}$Harbin Institute of Technology (Shenzhen) \\
$^{2}$Nanjing University, \
$^{3}$University of Science and Technology of China, \
$^{4}$Dexmal
}

\usepackage{graphicx}
\usepackage{multirow}
\usepackage{amsmath}
\usepackage{amssymb}
\usepackage{float}
\begin{document}

\maketitle
\thispagestyle{empty}
\pagestyle{empty}


\begin{abstract}
Vision-Language-Action (VLA) models leverage pretrained vision-language models (VLMs) to couple perception with robotic control, offering a promising path toward general-purpose embodied intelligence. 
However, current SOTA VLAs are primarily pretrained on multimodal tasks with limited relevance to embodied scenarios, and then finetuned to map explicit instructions to actions.
Consequently, due to the lack of reasoning-intensive pretraining and reasoning-guided manipulation, these models are unable to perform implicit human intention reasoning required for complex, real-world interactions.
To overcome these limitations, we propose \textbf{IntentionVLA}, a VLA framework with a curriculum training paradigm and an efficient inference mechanism. 
Our proposed method first leverages carefully designed reasoning data that combine intention inference, spatial grounding, and compact embodied reasoning, endowing the model with both reasoning and perception capabilities. In the following finetuning stage,  IntentionVLA employs the compact reasoning outputs as contextual guidance for action generation, enabling  fast inference under indirect instructions.
Experimental results show that IntentionVLA substantially outperforms $\pi_0$, achieving 18\% higher success rates with direct instructions and 28\% higher than ECoT under intention instructions.
On out-of-distribution intention tasks, IntentionVLA achieves over twice the success rate of all baselines, and further enables zero-shot human-robot interaction with 40\% success rate.
These results highlight IntentionVLA as a promising paradigm for next-generation human-robot interaction (HRI) systems.

\end{abstract}
\section{Introduction}

Recent advances in large-scale Vision–Language Models (VLMs)~\cite{llava, alayrac2022flamingo, bai2025qwen2.5vl, beyer2024paligemma} have demonstrated impressive multimodal understanding and reasoning capabilities. Building on this progress, Vision-Language-Action (VLA) models~\cite{brohan2023rt-1, OpenVLA, pi_0, li2024cogact, wen2025rosa}, adapt pretrained VLMs with manipulation data and have achieved remarkable performance in robotic control. However, most existing VLAs~\cite{OpenVLA, pi_0, li2024cogact, pertsch2025fast, wen2025tinyvla, liu2025hybridvla} are designed to directly predict actions, relying solely on action-only finetuning data and overfitting to low-level control signals. This paradigm severely weakens the intrinsic multimodal reasoning capabilities of pretrained VLMs, leading to limited contextual understanding and poor intent interpretation. As a result, such VLAs often struggle to comprehend human intentions then fail to execute accurately in cluttered real-world environments, which significantly restricts their applicability in human-robot interaction scenarios.

\begin{figure}[t] \centering
    \includegraphics[width=1 \linewidth]{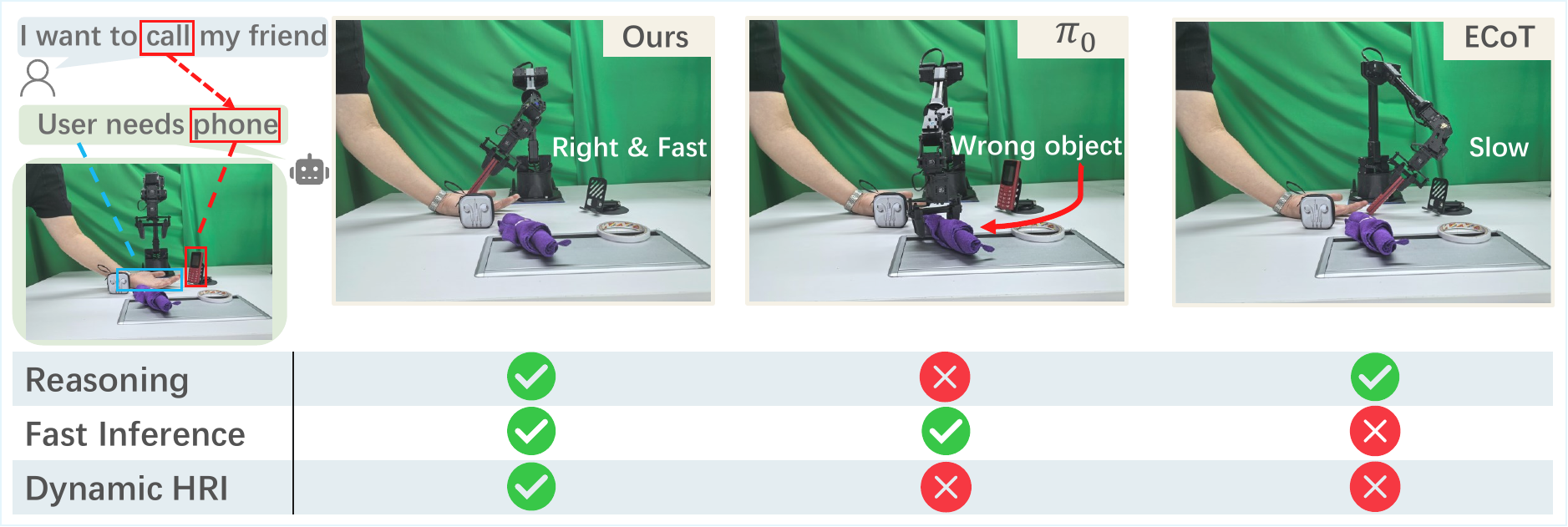}
    \caption{\textbf{Problems with existing VLAs}. Given the instruction ``I want to call my friend'',  ECoT (right) reaches for the phone but infers slowly, while $\pi_0$ (middle) misinterprets the instruction and grasps the rag. In contrast, our method (left) correctly infers the user's intention and enables rapid task completion.}
    \label{fig:intro_compare}
    \vspace{-1.0em}
\end{figure}

Given these limitations of current VLAs, we are motivated to develop a VLA model that can accurately interpret human intentions and perform correct actions in complex physical environments, as shown in Fig.~\ref{fig:intro_compare}. To achieve this, we face two critical challenges. First, existing manipulation datasets lack sufficient intention-aware annotations and contextual reasoning signals, making it difficult for models to learn the causal links between human goals, environmental states, and corresponding actions. Second, current approaches lack an explicit and lightweight semantic reasoning mechanism, making it hard to directly integrate the reasoning capabilities of VLMs into action generation.

In order to tackle these challenges, we propose \textbf{IntentionVLA}, a novel VLA framework pretrained with embodied intention reasoning supervision, also achieves accurate and fast inference. To address the lack of intention reasoning annotations during pretraining, we design a high-efficiency annotation pipeline that processes data collected from daily work environment. The annotations are represented in three complementary formats: intention reasoning, spatial reasoning, and compact reasoning, together forming a complete instruction-intention-grounding-action chain for model training. To bridge reasoning with action generation, IntentionVLA adopts a two-stage training paradigm: in the first stage, the model gains reasoning and perception capabilities from annotated data, while in the second stage, high-level reasoning is distilled into compact cues that guide a diffusion-based action generator, thereby explicitly enabling accurate action execution.

To evaluate the effectiveness of IntentionVLA, we conduct extensive real-world experiments covering direct instructions, diverse intention-driven tasks, and out-of-distribution manipulation scenarios. The results show that intentionVLA achieves strong performance and generalization, highlighting the benefits of our curriculum training paradigm and efficient inference mechanism. In summary, our main contributions are as follows:
\begin{itemize}

\item We present IntentionVLA, a unified VLA model trained on carefully curated intention reasoning data annotated by an efficient pipeline, and optimized through a two-stage paradigm with an efficient reasoning mechanism, which explicitly bridge high-level embodied intention reasoning with low-level action execution.

\item We conduct extensive experiments to demonstrate that IntentionVLA significantly outperforms state-of-the-art VLA baselines in all evaluation settings, exhibits strong generalization and real-time interaction capability.
\end{itemize}

\section{Related Work}
\textbf{Action-centric VLA Models}. Action-centric VLAs directly map multimodal inputs to low-level actions without reasoning. Autoregressive models such as RT-1~\cite{brohan2023rt-1}, OpenVLA~\cite{OpenVLA}, and $\pi_0$-FAST~\cite{pertsch2025fast} tokenize actions into discrete sequences, using next-token prediction as training objective to align with VLMs pretraining methodology. 
Diffusion-based approaches like  $\pi_0$~\cite{pi_0}, CogACT~\cite{li2024cogact}, and TinyVLA~\cite{wen2025tinyvla} generate continuous actions to reduce discretization error but at the cost of training efficiency. Hybrid designs attempt to balance the two paradigms. For example, HybridVLA~\cite{liu2025hybridvla} 
integrates diffusion denoising into the autoregressive next-token prediction process within a single LLM, aiming to exploit both the continuity of diffusion-based action generation and the intrinsic reasoning capabilities of autoregressive modeling.
However, finetuned solely on action trajectories, these models fail to generalize in intention reasoning intensive tasks that require inferring hidden user intentions. This highlights the need for models that go beyond direct action prediction toward explicit intention reasoning.

\textbf{Incorporating Reasoning into VLAs}. 
Research community has realized the limitations of action-centric VLAs, and several works enrich action policies with reasoning capabilities. 
Some~\cite{bu2024robodual} introduce implicit reasoning through latent actions, but the intermediate reasoning output is not interpretable.
Meanwhile, some works~\cite{yang2025instructvla, zhou2025chatvla} tailor model architecture for reasoning by introducing additional trainable modules such as mixture-of-experts, making the training process more sophisticated and computationally intensive.
One line of work~\cite{pi_0.5, cheang2025gr3technicalreport} attempts to train VLAs jointly on multimodal understanding data and manipulation data, but the substantial domain gap between generic multimodal tasks and embodied scenarios limits the effectiveness of this strategy, as large-scale multimodal pretraining brings only marginal improvements in embodied understanding.
Another line of work enriches manipulation datasets with structured data annotations while keeping the model architecture unchanged. ECoT~\cite{ecot} introduces embodied chain-of-thought reasoning, forcing the model to generate detailed textual plans before actions. Magma~\cite{Magma} leverages large-scale pre-training with surrogate objectives such as Set-of-Mark and Trace-of-Mark to ground reasoning in perception and action traces. While these approaches demonstrate improved interpretability and task decomposition, ECoT suffers from generating long textual reasoning chains, Magma needs an additional model to annotate input information, resulting in significant inference delays that hinder real-time responsiveness.
In contrast, \textbf{IntentionVLA} leverages diverse and effective reasoning data together with a simple but effective training paradigm, employing a streamlined reasoning mechanism, enabling robust and efficient intention inference that directly guides precise action generation.

\begin{figure}[tbp] \centering
    \includegraphics[width=0.96 \linewidth]{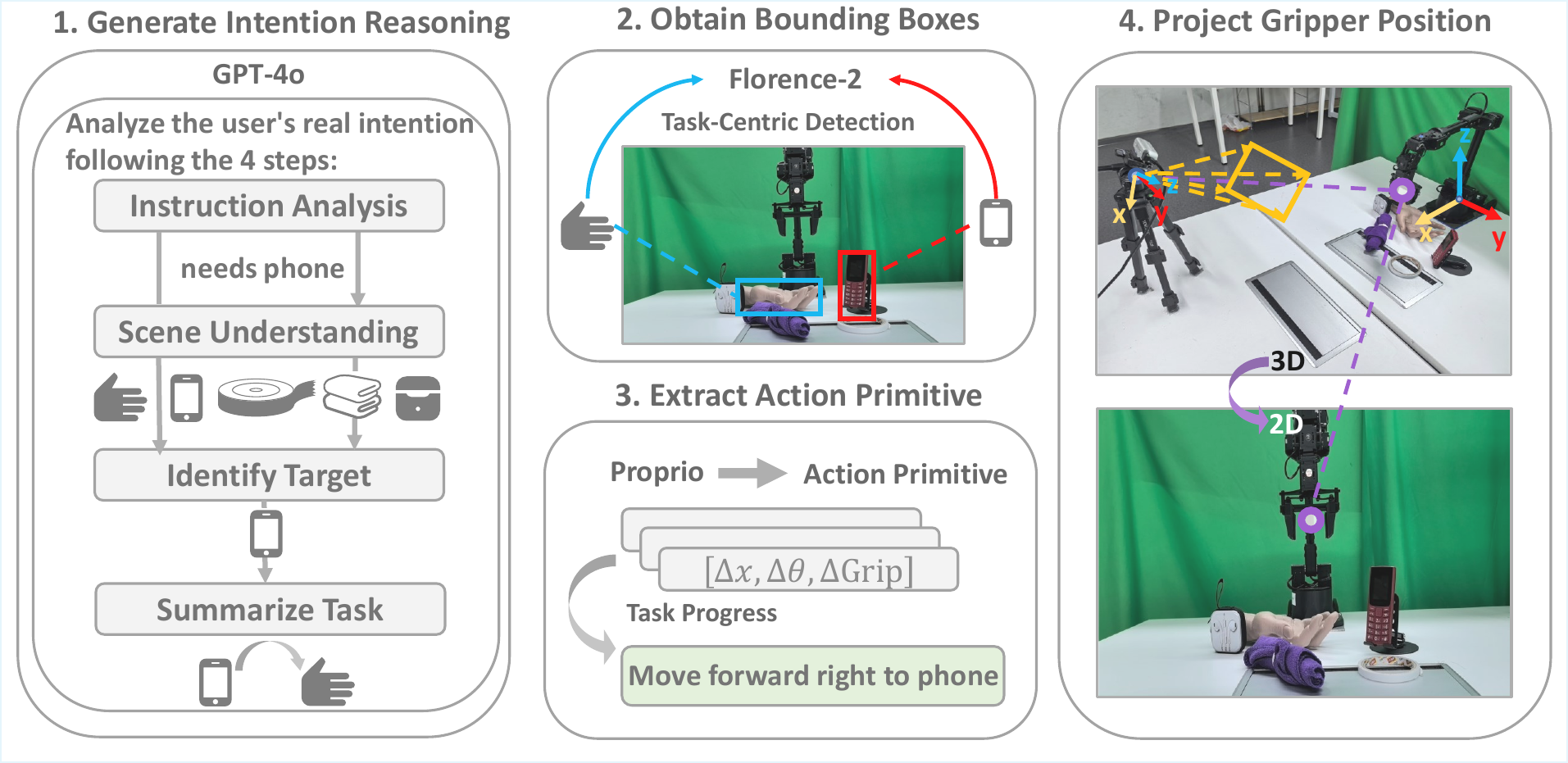}
    \caption{\textbf{Visualization of annotation module.} We visualize the four fully automated modules that constitute the data pipeline. The intermediate output of each module will be integrated to form the final reasoning data.}
    \label{fig:vis_anno_module}
    \vspace{-1.0em}
\end{figure}

\begin{figure*}[htbp] \centering
    \includegraphics[width=0.98 \linewidth]{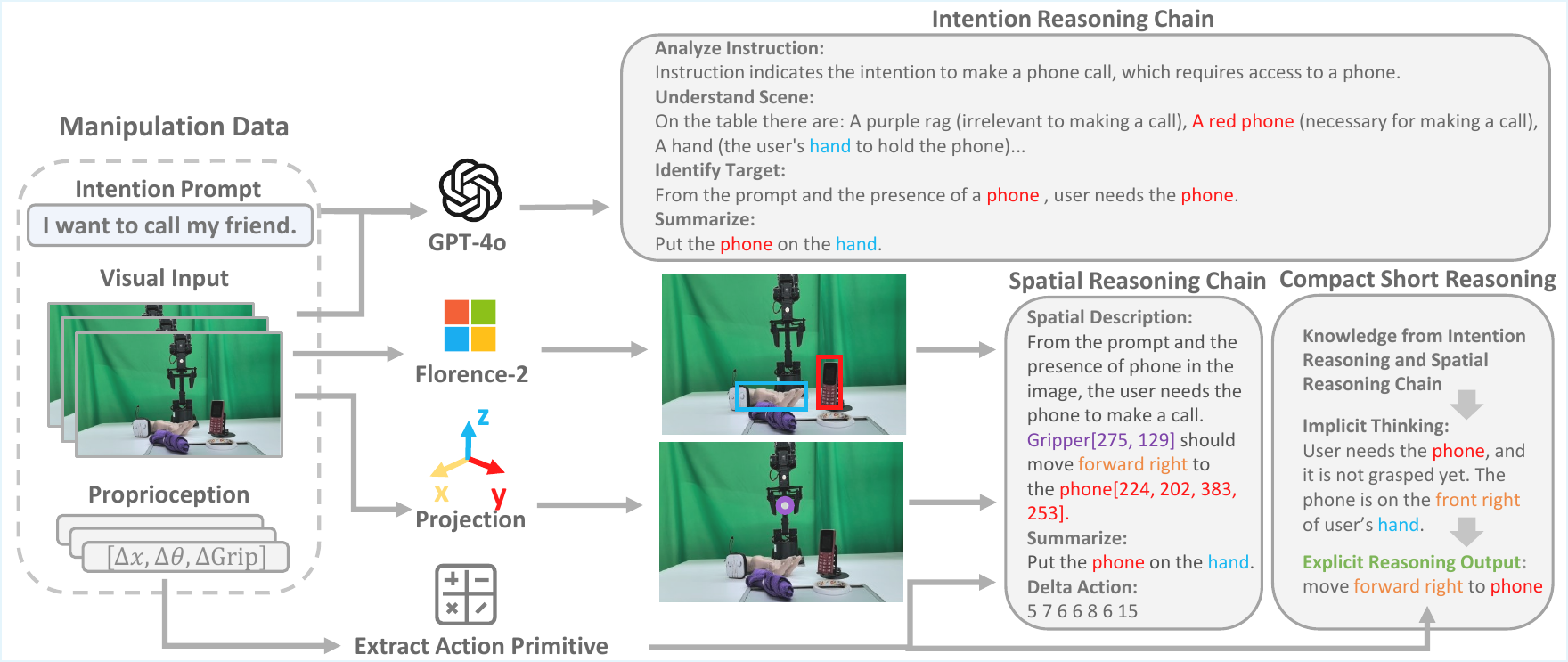}
    \caption{\textbf{Overview of our proposed reasoning data and efficient annotation pipeline.} The pipeline consists of 4 decoupled modules that can run in parallel for high annotation efficiency. Intention and spatial reasoning chains are further compressed into compact short reasoning for fast inference.}
    \label{fig:pipeline}
    \vspace{-1.0em}
\end{figure*}

\section{Method}
We propose IntentionVLA, a VLA framework that integrates embodied intention reasoning with effective action generation. 
Section A introduces the design of the embodied reasoning data format and provides a detailed description of the data annotation pipeline. 
Section B elaborates on the model architecture and highlights the advantages brought by its crafted design.
Section C focuses on the novel two-stage training paradigm, illustrating how embodied reasoning data can be leveraged efficiently to bridge high-level semantic reasoning with low-level action execution.

\subsection{Embodied Intention Reasoning Data Construction}
\textbf{Designing Embodied Intention Reasoning Data}.
While prior works~\cite{ecot,sun-etal-2025-emma-X} have explored embodied reasoning mainly as task planning from explicit instructions, we aim to extend it toward \textbf{intention inference}, enabling the model to interpret ambiguous instructions and ground hidden human intentions in cluttered environments for robust HRI. 
The overview of reasoning data and annotation pipeline are provided in Fig.~\ref{fig:pipeline}. We construct the \textbf{Embodied Intention Reasoning Dataset}, organized as a progressive curriculum with three objectives:
(1) Equip the model with the ability to infer user intentions in the embodied scenarios.
(2) Align inferred user intentions with visual observations, perceiving spatial relationships between the manipulator (end-effector) and task-relevant objects.
(3) Compress embodied intention reasoning into short textual sequences, enabling efficient short reasoning to explicitly guide action generation.

\textbf{Efficient Reasoning Data Annotation Pipeline}.
We design an efficient annotation pipeline that leverages pretrained vision–language models to automatically generate rich embodied reasoning signals, providing a scalable approach to align high-level intentions with low-level actions. The detailed visualization of each module is shown in Fig.~\ref{fig:vis_anno_module}. 

\textbf{Intention reasoning chain. }Given a user instruction and corresponding visual observation, we prompt GPT-4o~\cite{openai2024gpt4ocard} to decompose the task into 4 reasoning steps, producing a step-by-step analysis of the user’s latent intention. This yields an explicit intention reasoning chain that aligns natural language instructions with task goals.

\textbf{Spatial reasoning chain. }To bridge high-level intentions with visual grounding, we compress the intention chain into a one-sentence summary and enrich it with spatial information.
Specifically, we project the 3D end-effector position into the 2D image plane using calibrated camera parameters, providing more precise localization than detector-based baselines such as ECoT~\cite{ecot}. 
The target object’s bounding box is obtained via Florence-2~\cite{xiao2024florence}, a strong vision foundation model. For frames with invalid detections, we interpolate from the nearest valid frame to ensure temporal consistency. The 2D end-effector position and object bounding box are seamlessly embedded into the textual description, tightly aligning reasoning with visual observations. Additionally, we discretize end-effector pose deltas into 16 bins to form delta action tokens, serving as supervision for action generation. Integrating compressed intention reasoning, spatial grounding, and delta actions produces the spatial reasoning chain.

\textbf{Compact short reasoning. }To enable efficient inference, we further design a concise reasoning format in the form ``move \textless direction\textgreater\ to \textless object\textgreater''. The movement direction is derived from proprioceptive changes in the end-effector, while the object is dynamically determined by task progress. Specifically, we coarsely categorize task states into ``not yet grasped'' and ``already grasped,'' distinguished by gripper status. In the first case, the gripper moves toward the target object; in the latter, it moves toward the placement goal (e.g., the human hand). This compact reasoning effectively integrates task progress estimation and low-level motion planning, distilling knowledge from the richer reasoning chains into a format suitable for real-time execution.

By integrating these three reasoning formats, our pipeline produces rich, well-aligned multimodal supervision for embodied intention reasoning in daily work environments.

\begin{figure*}[htbp] \centering
    \includegraphics[width=0.98 \linewidth]{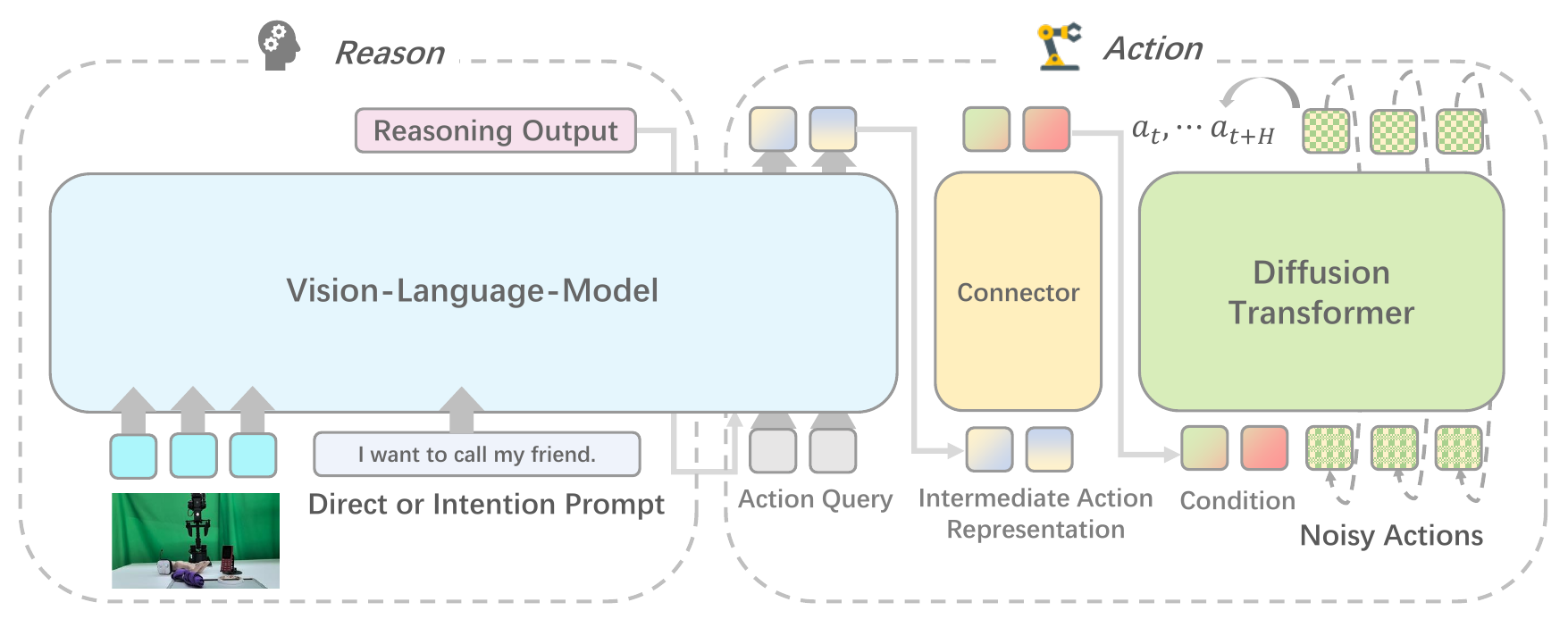}
    \caption{\textbf{Overview of IntentionVLA framework.} IntentionVLA achieves intention inference and reasoning-guided manipulation in one unified model. We first pretrain the VLM backbone with diverse intention reasoning data. Then we finetune the action module to decode action chunk which follows the compact reasoning output.}
    \label{fig:model}
    \vspace{-1.0em}
\end{figure*}

\subsection{Model Architecture}
We design IntentionVLA as a unified end-to-end framework that jointly performs embodied intention reasoning and action generation for manipulation tasks. Fig.~\ref{fig:model} shows the overall architecture, which builds upon a strong vision-language backbone while incorporating bridging modules that connect high-level semantic reasoning with low-level embodied actions.

\textbf{VLM Backbone for Embodied Intention Reasoning.}
At the core of our model is the Qwen2.5-7B~\cite{bai2025qwen2.5vl} backbone, which processes multimodal embodied inputs consisting of visual observations, textual instructions. 
We denote the single forward pass of the VLM as $f_{\mathrm{vlm}}$, and the autoregressive modeling process of the VLM as $f_{\mathrm{vlm}}^{\ell}$.
Unlike conventional VLA models that are limited to action-only supervision, our backbone is trained with embodied intention reasoning data, enabling it to produce outputs in multiple forms: (1) text tokens reflecting inferred user intentions, (2) predictions of end-effector and goal positions, and (3) discrete action tokens. This unified representation empowers the model to reason about abstract user intent while grounding it in task-relevant embodied contexts.

\textbf{Intermediate Action Representations via Learnable Queries.}
To further extract intermediate action representations from the VLM, we append a set of learnable queries 
$Q \in \mathbb{R}^{N \times D}$ to the multimodal input sequence, where $N$ is the amount of query tokens, $D$ is the VLM's hidden dimension. These queries attend to the hidden states of the VLM, selectively capturing the semantic cues necessary for action generation. By decoupling intention reasoning from action representation, these queries serve as a crucial bridge, transforming abstract reasoning into action-oriented embeddings.

\textbf{Connector: Refining Intermediate Representations for Action Conditioning.}
The intermediate action representations 
$z \in \mathbb{R}^{N \times D}$ extracted by learnable queries remain relatively abstract and require further refinement before being used in the diffusion denoising process. To this end, we introduce a 4-layer transformer~\cite{vaswani2017attention} module, termed the connector. We denote the forward process of connector as $f_{\mathrm{Con}}$ . This module refines intermediate action representations into well-structured action conditions 
$c \in \mathbb{R}^{N \times D_{diff}}$, effectively transferring information from the semantic reasoning domain of the VLM to the action domain of diffusion model. $D_{diff}$ is the hidden state size of diffusion model. By aligning abstract reasoning outcomes with the spatial-temporal requirements of manipulation tasks, the connector ensures that intention understanding can be transformed into precise, executable guidance for action generation.

\textbf{Action Generation with Diffusion Transformer.}
Finally, the connector outputs are concatenated with Gaussian noise vectors and fed into a Denoising Transformer (DiT)~\cite{dit}, which performs self-attention denoising to generate robust, condition-aware action trajectories. A lightweight MLP head maps the denoised representations to the 7-DoF pose changes of the end-effector, including translational (x, y, z), rotational (r, p, y), and gripper states (open, close). This design enables the model to produce precise, executable action outputs while maintaining flexibility and scalability across diverse HRI scenarios.

\begin{table*}[!t]
\centering
\caption{Performance comparison on in-distribution tasks with WidowX under different instructions.}
\resizebox{\linewidth}{!}{
\begin{tabular}{lcccccccc}
\hline
\multirow{2}{*}{\textbf{Task}} & \multicolumn{4}{c}{\textbf{Direct Instruction}} & \multicolumn{4}{c}{\textbf{Intention Instruction}} \\
\cline{2-5} \cline{6-9}
 & \textbf{CogACT} & \textbf{\boldmath $\pi_0$}
 & \textbf{ECoT} & \textbf{IntentionVLA} & \textbf{CogACT} & \textbf{\boldmath $\pi_0$}
 & \textbf{ECoT} & \textbf{IntentionVLA} \\
\hline
Phone on hand          & 40    & 40    & 20    & 60    & 0     & 10    & 10    & 70    \\
Gluestick on hand      & 40    & 10    & 30    & 30    & 20    & 10    & 30    & 40    \\
Marker on hand         & 50    & 20    & 20    & 40    & 0     & 20    & 10    & 50    \\
Phone on charger       & 40    & 50    & 20    & 90    & 10    & 40    & 20    & 30    \\
Rag on hand            & 50    & 40    & 30    & 40    & 10    & 30    & 20    & 30    \\
Pencil box on hand     & 30    & 20    & 10    & 30    & 0     & 10    & 10    & 50    \\
\hline
\textbf{Average Success Rate}   & 
41.7  & 30     & 21.7  & \textbf{48.3}  & 6.7   & 20     & 16.7  & \textbf{45}    \\
\hline
\end{tabular}
}
\label{tab:success_rates_compare}
\end{table*}

\textbf{Compact Reasoning Guided Inference Time Scaling.}
To enable low-latency decision-making, IntentionVLA performs compact reasoning at inference time. Instead of generating long reasoning chains, the  model first autoregressively produces a short reasoning sequence (e.g., ``move foward right to phone'') within 0.2s. This sequence encapsulates both user intention and task progress estimation, providing a minimal yet sufficient embodied context. Conditioned on the original input and this compact reasoning, the VLA generates the intermediate action representation in a single forward pass. This design achieves real-time responsiveness while maintaining accurate intention grounding and robust action execution in cluttered HRI scenarios. 
The whole inference procedure can be formulated as follow:

\begin{equation}
\mathbf{c} \;=\; f_{\mathrm{Con}}\!\Big( f_{\mathrm{vlm}}\big([\mathbf{o}_t;\;\mathbf{\ell};\; f_{vlm}^\ell(\mathbf{o}_t,\mathbf{\ell});\; \mathbf{Q}\;]\big) \Big)
\label{eq:condition}
\end{equation}


\begin{equation}
\hat{\boldsymbol{\epsilon}} \;=\; \boldsymbol{\epsilon}_\theta\!\big(\; \mathbf{A}_{k},\; k,\; \mathbf{c} \;\big)
\label{eq:predict_noise}
\end{equation}


\begin{equation} \widehat{\mathbf{A}}_0 \;=\; \frac{\mathbf{A}_k - \sqrt{1-\bar\alpha_k}\,\hat{\boldsymbol{\epsilon}}} {\sqrt{\bar\alpha_k}} \label{eq:ddim_reconstruct} \end{equation}
where $\mathbf{o}_t$ is the current observation and $\ell$ is the language instruction. $k$ is the diffusion timestep, ${\mathbf{A}}_{k}$ is noisy action chunk at diffusion timestep $k$. $\boldsymbol{\epsilon}_\theta$ stands for the DiT network that predicts the noise $\hat{\boldsymbol{\epsilon}}$ added to the action, conditioned on timestep $k$ and condition $c$. $\bar\alpha_k$ is a hyperparameter defined by the noise scheduler.

During inference, we start from Gaussian noise $\mathbf{A}_T \sim \mathcal{N}(0, I)$, and iteratively denoise it using the DiT network $\boldsymbol{\epsilon}_\theta$, following the (\ref{eq:ddim_reconstruct}), to obtain the estimate of clean action chunk $\hat{\mathbf{A}}_0$.

\subsection{Two-Stage Training Scheme}

To enable IntentionVLA to effectively integrate high-level embodied reasoning with low-level action generation, we adopt a two-stage training scheme. This curriculum-based design ensures that the model first acquires robust multimodal reasoning and perception skills before being extended to the action space, fully exploiting the carefully designed reasoning formats.

\textbf{Stage 1: Embodied Intention Reasoning and Spatial Perception.}
In the first stage, training focuses exclusively on the VLM backbone. The objective is to equip the VLM with embodied intention reasoning and spatial perception capability. Specifically, the model is trained to infer abstract user intentions while simultaneously localizing key elements of manipulation tasks, such as the end-effector and target object positions within the visual scene.
During this stage, the model outputs either textual intention tokens or discrete action tokens. 
The training objective of this stage is to minimize:
\begin{equation}
\begin{aligned}
\mathcal{L}_{\text{CE}} = \mathbb{E}_{\mathcal{D}} \left[ H(x_{1:M}, f_{vlm}^\ell(\mathbf{o}_t, \ell))
\right]
\end{aligned}
\end{equation}
where $H(x_{1:M}, f_{vlm}^\ell(\mathbf{o}_t, \ell)$ is the cross entropy between the input tokens $x_{1:M} \in \mathcal{D}$ and predicted logits $f_{vlm}^\ell(\mathbf{o}_t, \ell)$. This learning objective allows the VLM to align multimodal observations with appropriate reasoning outcomes. By the end of this stage, the VLM acquires strong embodied reasoning skills that form the foundation for downstream action generation.

\textbf{Stage 2: Reasoning-guided Action Query Learning and Query-Conditioned Generation.}
In the second stage, we train the modules responsible for translating semantic reasoning into executable actions: the learnable action queries, the connector Transformer, and the DiT-based action generator. DiT uses the output $c$ from connector as the condition. Diffusion loss based on mean squared error (MSE) is employed to supervise the denoising process. 

\begin{equation}
\mathcal{L}_{\text{Diff}} = \mathbb{E}_{\mathbf{A}_0, k, \boldsymbol{\epsilon}} \left[ \left\| \boldsymbol{\epsilon} - \boldsymbol{\epsilon}_\theta(\mathbf{A}_k, k, c) \right\|^2 \right]
\end{equation}
where $\mathbf{A}_0 \in \Gamma$ is action chunk in manipulation dataset, $\mathbf{A}_k$ is the noisy action chunk, $\epsilon \sim \mathcal{N}(0, I)$ is the sampled noise.
Action queries learn to extract contextual information from the compact reasoning. Importantly, the action queries are produced by the VLM but detached from the computation graph before being passed to the connector. This design prevents gradients from flowing back into the VLM during this stage, ensuring that the reasoning and perception capabilities learned in Stage 1 remain intact.



\section{Experiment}

\begin{figure}[tbp] \centering
    \includegraphics[width=0.98 \linewidth]{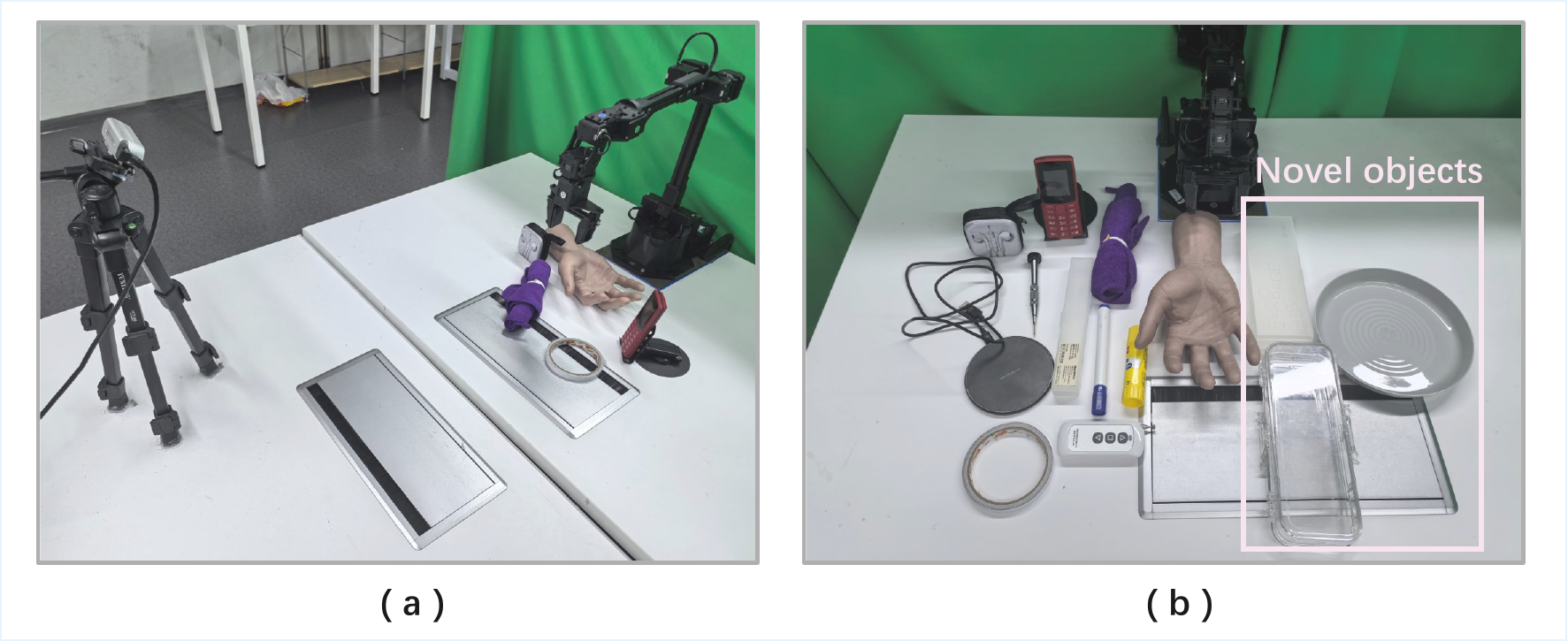}
    \caption{\textbf{Real-world experiment setting}}
    \label{fig:real-world setup}
    \vspace{-1.0em}
\end{figure}

\begin{figure*}[htbp] \centering
    \includegraphics[width=0.98 \linewidth]{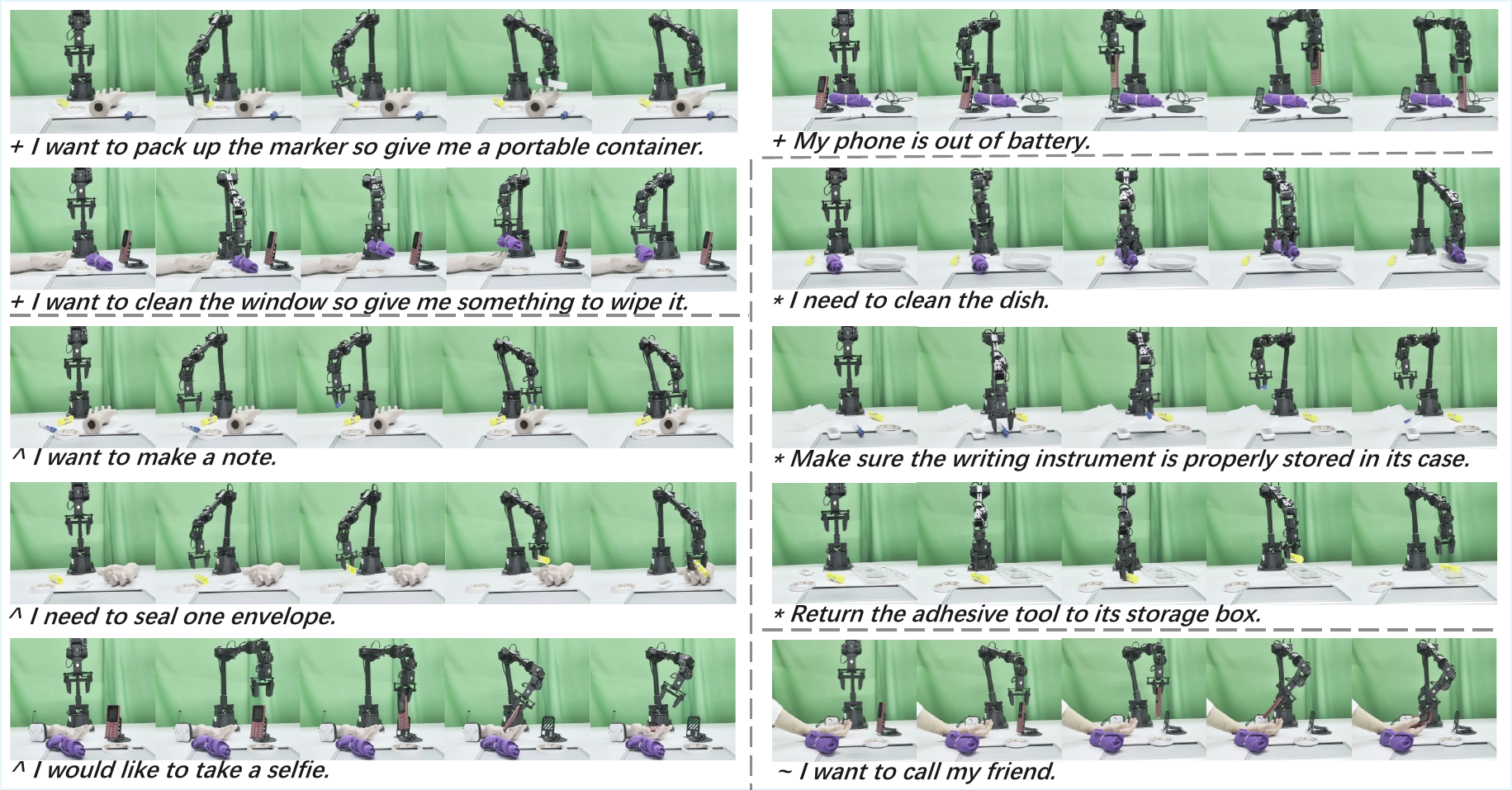}
    \caption{\textbf{Qualitative results of IntentionVLA on different task settings.}
    We present four kinds of task here: where
    $+$ denotes in-distribution intention task, 
    \textasciicircum\ denotes unseen instruction task, 
    $*$ denotes novel object manipulation, 
    \textasciitilde\ denotes human-robot interaction task.}
    \label{fig:vis_rollouts} 
\end{figure*}

We evaluate IntentionVLA and baseline models on in-distribution (ID), out-of-distribution (OOD), and zero-shot human–robot interaction (HRI) tasks. 
ID tasks covers direct and intention instructions, OOD tasks tests generalization to unseen instructions and novel objects, both using a hand model. Zero-shot HRI instead employs real human hands to validate performance in real-world HRI scenarios.
We argue that intention instructions are quite challenging for these baseline models, as training data for all 6 in-distribution tasks share similar objects shown in Fig.~\ref{fig:real-world setup}. There are five objects in each evaluation scene, and only two of them are implied by the instruction, the others are all visual distractors. So only one model correctly interprets the human intention and grounding the corresponding targets can it complete the task.

IntentionVLA consistently achieves superior performance across all evaluation settings, and qualitative results are shown in Fig.~\ref{fig:vis_rollouts}.

\subsection{Evaluation Metric and Experiment Setup}


\textbf{Evaluation Metric:}
We adopt task success rate as the evaluation metric, measured over 10 trials for each task.
To rigorously assess HRI capability, target object poses and placement poses are randomized. A trial is considered successful only if the robot completes the full interaction correctly. For example, in the Phone on hand task, the robot must securely grasp the phone, reorient it, and stably place it in a position that allows comfortable human grasp. Any failure due to mis-grasp, slippage during execution, or unstable placement is counted as unsuccessful.

\textbf{Experimental Setup:}
We conduct evaluation in a realistic office environment as shown in Fig.~\ref{fig:real-world setup}~(a). The robotic platform is a WidowX-250s robotic arm equipped with a Realsense D435i camera for visual perception. 
IntentionVLA is trained with the proposed two-stage training recipe, reasoning baseline ECoT~\cite{ecot} is trained on the same data but in embodied chain-of-thought format, whereas action-centirc baseline CogAct~\cite{li2024cogact} and $\pi_0$~\cite{pi_0} are supervised to directly predict action chunk with diffusion or flow-matching loss.





\subsection{Intention Reasoning in In-Distribution Tasks}
We first compare IntentionVLA with strong baselines on direct and intention instruction tasks. We evaluate 6 in-distribution tasks in total, each task is conducted on direct instructions for 10 trails and intention instructions for another 10 trails. Take the task Phone on charger as an example, the direct instruction is ``Put the phone on the charger'' while the intention instruction is ``My phone is out of battery''. 

The detailed result is provided in TABLE.~\ref{tab:success_rates_compare}. Under direct instructions, IntentionVLA achieves the highest average success rate of 48.3\%, outperforming the reasoning baseline ECoT by 26.6\% and the action baseline $\pi_0$ by 18.3\%. 
In terms of intention instructions, where the model must infer implicit user goals rather than follow explicit commands, IntentionVLA maintains robust performance and gets 45\%  average success rate, while baselines drop sharply, all lower than 20\%. 
Although all baselines were trained with intention instruction data, their training paradigms emphasize either direct instruction-action mapping (CogACT, $\pi_0$) or long embodied reasoning chains (ECoT), both of which fail to capture implicit intention reasoning. 
Notably, CogACT achieves only 6.7\% on average, hardly handling intention-driven tasks, and is thus excluded from subsequent evaluations.

\subsection{Generalization to OOD setting and Real-World HRI}

\begin{figure}[bp] \centering
    \vspace{-1.0em}

    \includegraphics[width=0.98 \linewidth]{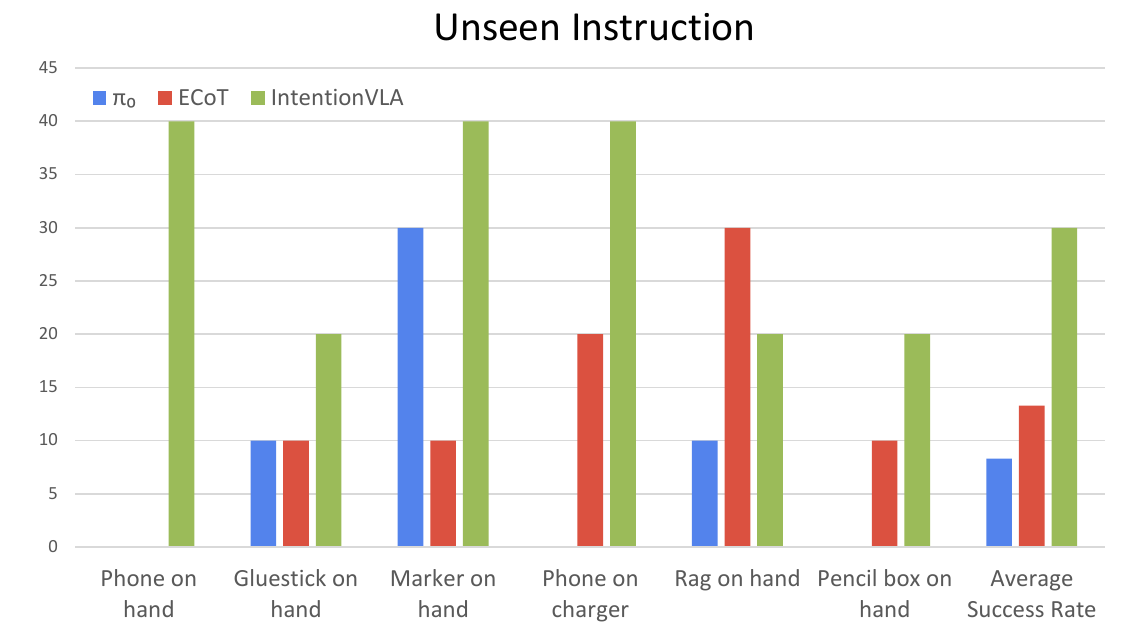}
    \caption{Performance on unseen instructions.}
    \label{fig:unseen_instruct_performance}
\end{figure}

\begin{table}[tbp]
\centering
\caption{Performance on novel objects manipulation task.}
\resizebox{0.5\textwidth}{!}{
\begin{tabular}{lccc}
\hline
\textbf{Model} & \textbf{Marker in pencil box} & \textbf{Rag on plate} & \textbf{Gluestick in storage box}\\
\hline
$\pi_0$ & 0 & 20 & 0\\
ECoT & 0 & 20 & 10\\
IntentionVLA & \textbf{20} & \textbf{30} & \textbf{10}\\

\hline
\end{tabular}}
\label{tab:novel_objects_task}
\vspace{-1.0em}
\end{table}

\begin{table}[tbp]
\centering
\caption{Performance on HRI Task}
\resizebox{0.5\textwidth}{!}{
\begin{tabular}{lccc}
\hline
\textbf{Model} & \textbf{Phone on real hand} & \textbf{Phone on
moving hand} & \textbf{Average inference time} \\
\hline
$\pi_0$ & 0 & 0 & 0.08 \\
ECoT & 20 & 0 & 3.41\\
IntentionVLA & \textbf{40} & \textbf{30} & 0.72\\
\hline
\end{tabular}}
\label{tab:HRI-task}
\vspace{-1.0em}
\end{table}
To rigorously evaluate IntentionVLA’s robustness and generalization, we design three progressively challenging evaluation settings that push the model to its operational limits.

\textbf{Unseen instructions.}
\begin{table*}[htbp]
\centering
\caption{Performance comparison on multimodal understanding and VQA benchmarks. * indicates multimodal large language models, and the rest are Vision-Language-Action (VLA) models.}
\resizebox{0.9\linewidth}{!}{
    \begin{tabular}{lc|cccc|cc}
    \hline
    \multirow{2}{*}{Method} & \multirow{2}{*}{Params} & \multicolumn{4}{c|}{\textbf{Multimodal Understanding Benchmarks}} & \multicolumn{2}{c}{\textbf{VQA Benchmarks}} \\
\cline{3-8}          &       & \textbf{MMMU~\cite{yue2024mmmu}} & \textbf{MMStar~\cite{MMstar}} & \textbf{MME~\cite{fu2024mme}} & \textbf{HallBench~\cite{guan2024hallusionbench}} & \textbf{AI2D~\cite{AI2D}} & \textbf{RealWorldQA~\cite{team2024realworldqa}} \\
    \hline
    LLaVA*~\cite{llava} & 7B    & 27.1  & 34.1  & 1075.5 & 21.6  & 48.3  & 45.8 \\
    OpenVLA~\cite{OpenVLA} & 7B    & 0.0   & 0.0   & 0.0   & 0.0   & 0.0   & 0.0 \\
    ECoT~\cite{ecot} & 7B    & 16.2  & 19.1  & 0.0   & 3.1   & 0.0   & 29.8 \\
    \textbf{IntentionVLA(Ours)} & 7B    & \textbf{28.4} & \textbf{28.9} & \textbf{1659.8} & \textbf{51.6} & \textbf{32.9} & \textbf{36.9} \\
    \hline
    \end{tabular}%
}
\label{tab:multi_vqa}%
\end{table*}
\begin{table*}[htbp]
\centering
\caption{Ablations on pretraining data}
\resizebox{0.85\textwidth}{!}{
\begin{tabular}{lcccc}
\hline
\textbf{Task} & \textbf{Full data} & \textbf{w/o intention reason} & \textbf{w/o spatial reason} & \textbf{w/o compact reasoning} \\
\hline
Phone on hand          & 70 & 30 & 30 & 40 \\
Gluestick on hand      & 40 & 30 & 30 & 10 \\
Marker on hand         & 50 & 20 & 30 & 30 \\
Phone on charger       & 30 & 30 & 30 & 30 \\
Rag on hand            & 30 & 40 & 40 & 30 \\
Pencil box on hand     & 50 & 20 & 30 & 30 \\
\hline
\textbf{Average Success Rate} & \textbf{45} & 28.3 & 31.7 & 28.3 \\
\hline
\end{tabular}}
\label{tab:ab_pretrain_data}
\end{table*}
In this setting, we still evaluate on the same 6 in distribution tasks. However, for each task, we evaluate the model using 5 intention instructions that were not present in the training data. Each unseen instruction is tested with 2 rollouts, resulting in a total of 10 rollouts per task. This setting imposes higher demands on the VLA’s capacity for intention inference.
As illustrated by the bar chart in Fig.~\ref{fig:unseen_instruct_performance}, IntentionVLA consistently outperforms all baselines across nearly all tasks, reaching an average success rate of 30\%. ECoT, despite introducing reasoning before action, attains only 13.3\% sucees rate on average, less than half of IntentionVLA.
Performance degrades even further for $\pi_0$, which completely fails on three tasks (Phone on hand, Phone on charger, and Pencil box on hand) and yields only 8.3\% success overall,  it can barely execute right action under unseen intention instructions.
These results demonstrates that our reasoning data covering intention inference, spatial grounding, and compact embodied reasoning, collaboratively provides richer supervision than ECoT’s longer but less effective reasoning chains, enabling superior generalization to unseen intention formulations.

\textbf{Novel object manipulation.}
This setting further increases the difficulty beyond the unseen instruction setting. We design 3 new tasks in which the model is not only presented with unseen intention instructions, but also needs to interact with previously unseen objects during task execution. This requires the model to align the implicit task goals embedded in the novel instructions with the functional affordances of unfamiliar objects, which places extreme demands on the model’s generalizable reasoning, perceptual grounding, and reasoning-following manipulation capabilities. 
The three novel objects (big pencil box, grey plate, transparent storage box) are shown in Fig.~\ref{fig:real-world setup}~(b), and the corresponding task instructions are:
``Make sure the writing instrument is properly stored in its case.''
``I need to clean the dish.''
``Return the adhesive tool to its storage box.''

The result is provided in TABLE.~\ref{tab:novel_objects_task}.
Even under such difficult setting, IntentionVLA also generalizes effectively to all novel objects, including visually challenging transparent ones. IntentionVLA is the only model that can complete task Marker in pencil box. It consistently outperforms both $\pi_0$ and ECoT, which often collapse to 0\% success due to their limited perception-reasoning integration.

\textbf{Zero-shot HRI.}
Finally, we conduct human-in-the-loop evaluations to thoroughly assess the model’s real-time responsiveness. 
We adopt the ID task ``Phone on hand'' as the base scenario, providing the instruction: ``I want to call my friend.''
In the Phone on real hand condition, the human evaluator’s hand remains stationary throughout the trial. In contrast, in the Phone on moving hand, the human hand performs one random movement during the robot’s rollout, explicitly testing the model’s capacity for real-time adaptation and dynamic response to unexpected human motion.

The result is provided in TABLE.~\ref{tab:HRI-task}. $\pi_0$ fails completely, always grasping incorrect objects, while ECoT suffers from severe inference latency (over 5 times slower than IntentionVLA), making it infeasible to respond to dynamic human motion. Thanks to the efficient compact reasoning mechanism, IntentionVLA achieves 40\% and 30\% success rates on Phone on real hand and Phone on moving hand, respectively. These results demonstrate that IntentionVLA enables robust, intention-aware, and real-time HRI in dynamic scenarios.

\subsection{Results on Multimodal Understanding}
The above experiments demonstrate the strong perception and intention reasoning capabilities of IntentionVLA in real-world settings. To further assess its general multimodal perception and understanding ability, we evaluate the model using VLMEvalKit~\cite{duan2024vlmevalkit} on diverse multimodal understanding and VQA benchmarks. As shown in TABLE.~\ref{tab:multi_vqa}, IntentionVLA consistently surpasses all VLA baselines, benefiting from the diversity of our reasoning data and the effectiveness of the two-stage training paradigm. 
Notably, IntentionVLA outperforms the strong MLLM LLaVA~\cite{llava} on three benchmark, including MMMU~\cite{yue2024mmmu}, MME~\cite{fu2024mme}, and even achieving 51.6\% accuracy on HallBench~\cite{guan2024hallusionbench}. This highlights IntentionVLA’s low hallucination property and its suitability for accurate and safe real-time interaction in realistic scenarios.

\subsection{Ablation Studies}
We attribute IntentionVLA’s strong intention understanding and precise execution to two key factors: pretraining on effective reasoning data and finetuning with compact reasoning as efficient guidance. To validate this, we conduct ablations on both components.

\textbf{Effects of pretraining data.} 
As shown in TABLE.~\ref{tab:ab_pretrain_data}, removing intention reasoning or spatial reasoning reduces success rates by 16.7\% and 13.3\% respectively, highlighting their roles in human intention inference and intention–action grounding. Most critically, incorporating compact reasoning boosts performance from 28.3\% to 45\% (a 59\% relative gain), confirming it as an effective bridge between high-level reasoning and low-level control.

\begin{table}[tbp]
\centering
\caption{Ablations on finetuning strategy.}
\resizebox{0.45\textwidth}{!}{
\begin{tabular}{lcc}
\hline
\textbf{Task} & \textbf{VLM \& action expert} & \textbf{Only action expert} \\
\hline
Phone on hand        & 40 & 70 \\
Gluestick on hand    & 20 & 40 \\
Marker on hand       & 70 & 50 \\
Phone on charger     & 50 & 30 \\
Rag on hand          & 40 & 30 \\
Pencil box on hand   & 30 & 50 \\
\hline
\textbf{Average Success Rate} & 41.7 & \textbf{45} \\
\hline
\end{tabular}}
\label{tab:ab_finetune}
\vspace{-1.0em}
\end{table}

\textbf{Ablations on finetuning strategy.} 
We further compare whether to jointly optimize the VLM and action expert or to finetune only the action expert with frozen VLM. The result is provided in TABLE.~\ref{tab:ab_finetune}. While both yield similar average success rates, the latter one produces more stable rotations and smoother trajectories. Hence, Only action expert is adopted as the default strategy, and all reported results of IntentionVLA follow this setting.

   
\section{Conclusions}
We present IntentionVLA, which advances VLA training with a novel intention reasoning-driven paradigm and an efficient compact reasoning mechanism. By pretraining on diverse reasoning data and transferring these capabilities to action finetuning, the model acquires generalizable intention understanding. During inference, compact reasoning enables precise action guidance while ensuring real-time efficiency. Extensive experiments confirm IntentionVLA's superior generalization to unseen instructions and novel objects, as well as responsive human-robot interaction, highlighting the effectiveness of our training strategy and reasoning design.


\addtolength{\textheight}{-12cm}   









\bibliographystyle{IEEEtranBST/IEEEtran}
\bibliography{IEEEtranBST/ref}

\begin{thebibliography}{10}
\providecommand{\url}[1]{#1}
\csname url@rmstyle\endcsname
\providecommand{\newblock}{\relax}
\providecommand{\bibinfo}[2]{#2}
\providecommand\BIBentrySTDinterwordspacing{\spaceskip=0pt\relax}
\providecommand\BIBentryALTinterwordstretchfactor{4}
\providecommand\BIBentryALTinterwordspacing{\spaceskip=\fontdimen2\font plus
\BIBentryALTinterwordstretchfactor\fontdimen3\font minus \fontdimen4\font\relax}
\providecommand\BIBforeignlanguage[2]{{%
\expandafter\ifx\csname l@#1\endcsname\relax
\typeout{** WARNING: IEEEtran.bst: No hyphenation pattern has been}%
\typeout{** loaded for the language `#1'. Using the pattern for}%
\typeout{** the default language instead.}%
\else
\language=\csname l@#1\endcsname
\fi
#2}}

\bibitem{llava}
H.~Liu, C.~Li, Q.~Wu, \emph{et~al.}, ``Visual instruction tuning,'' \emph{Advances in neural information processing systems}, vol.~36, pp. 34\,892--34\,916, 2023.

\bibitem{alayrac2022flamingo}
J.-B. Alayrac, J.~Donahue, P.~Luc, \emph{et~al.}, ``Flamingo: a visual language model for few-shot learning,'' \emph{Advances in neural information processing systems}, vol.~35, pp. 23\,716--23\,736, 2022.

\bibitem{bai2025qwen2.5vl}
S.~Bai, K.~Chen, X.~Liu, \emph{et~al.}, ``Qwen2. 5-vl technical report,'' \emph{arXiv preprint arXiv:2502.13923}, 2025.

\bibitem{beyer2024paligemma}
L.~Beyer, A.~Steiner, A.~S. Pinto, \emph{et~al.}, ``Paligemma: A versatile 3b vlm for transfer,'' \emph{arXiv preprint arXiv:2407.07726}, 2024.

\bibitem{brohan2023rt-1}
A.~Brohan, N.~Brown, J.~Carbajal, \emph{et~al.}, ``Rt-1: Robotics {Transformer} for {Real}-{World} {Control} at {Scale}.'' in \emph{Robotics: Science and {Systems} {Conference} ({RSS})}, 2023.

\bibitem{OpenVLA}
M.~J. Kim, K.~Pertsch, S.~Karamcheti, \emph{et~al.}, ``Openvla: An open-source vision-language-action model,'' in \emph{Proceedings of The 8th Conference on Robot Learning}, ser. Proceedings of Machine Learning Research, P.~Agrawal, O.~Kroemer, and W.~Burgard, Eds., vol. 270.\hskip 1em plus 0.5em minus 0.4em\relax PMLR, 06--09 Nov 2025, pp. 2679--2713.

\bibitem{pi_0}
K.~Black, N.~Brown, D.~Driess, \emph{et~al.}, ``$\pi{}0$: A {Vision}-{Language}-{Action} {Flow} {Model} for {General} {Robot} {Control},'' in \emph{Robotics: Science and {Systems} {XXI}}, vol. abs/2410.24164.\hskip 1em plus 0.5em minus 0.4em\relax {Robotics: Science and Systems Foundation}, 2025.

\bibitem{li2024cogact}
Q.~Li, Y.~Liang, Z.~Wang, \emph{et~al.}, ``Cogact: A foundational vision-language-action model for synergizing cognition and action in robotic manipulation,'' \emph{arXiv preprint arXiv:2411.19650}, 2024.

\bibitem{wen2025rosa}
Y.~Wen, K.~Gu, H.~Liu, \emph{et~al.}, ``Rosa: Harnessing robot states for vision-language and action alignment,'' \emph{arXiv preprint arXiv:2506.13679}, 2025.

\bibitem{pertsch2025fast}
K.~Pertsch, K.~Stachowicz, B.~Ichter, \emph{et~al.}, ``Fast: Efficient action tokenization for vision-language-action models,'' \emph{arXiv preprint arXiv:2501.09747}, 2025.

\bibitem{wen2025tinyvla}
J.~Wen, Y.~Zhu, J.~Li, \emph{et~al.}, ``Tinyvla: Towards fast, data-efficient vision-language-action models for robotic manipulation,'' \emph{IEEE Robotics and Automation Letters}, 2025.

\bibitem{liu2025hybridvla}
J.~Liu, H.~Chen, P.~An, \emph{et~al.}, ``Hybridvla: Collaborative diffusion and autoregression in a unified vision-language-action model,'' 2025.

\bibitem{bu2024robodual}
Q.~Bu, H.~Li, L.~Chen, \emph{et~al.}, ``Towards synergistic, generalized, and efficient dual-system for robotic manipulation,'' \emph{arXiv preprint arXiv:2410.08001}, 2024.

\bibitem{yang2025instructvla}
S.~Yang, H.~Li, Y.~Chen, \emph{et~al.}, ``Instructvla: Vision-language-action instruction tuning from understanding to manipulation,'' \emph{arXiv preprint arXiv:2507.17520}, 2025.

\bibitem{zhou2025chatvla}
Z.~Zhou, Y.~Zhu, M.~Zhu, \emph{et~al.}, ``Chatvla: Unified multimodal understanding and robot control with vision-language-action model,'' 2025.

\bibitem{pi_0.5}
P.~Intelligence, K.~Black, N.~Brown, \emph{et~al.}, ``$\pi_{0.5}$: a vision-language-action model with open-world generalization,'' 2025.

\bibitem{cheang2025gr3technicalreport}
C.~Cheang, S.~Chen, Z.~Cui, \emph{et~al.}, ``Gr-3 technical report,'' 2025.

\bibitem{ecot}
M.~Zawalski, W.~Chen, K.~Pertsch, \emph{et~al.}, ``Robotic control via embodied chain-of-thought reasoning,'' in \emph{8th Annual Conference on Robot Learning}, 2024.

\bibitem{Magma}
J.~Yang, R.~Tan, Q.~Wu, \emph{et~al.}, ``Magma: A foundation model for multimodal ai agents,'' in \emph{Proceedings of the IEEE/CVF Conference on Computer Vision and Pattern Recognition (CVPR)}, June 2025, pp. 14\,203--14\,214.

\bibitem{sun-etal-2025-emma-X}
Q.~Sun, P.~Hong, T.~D. Pala, \emph{et~al.}, ``Emma-{X}: An embodied multimodal action model with grounded chain of thought and look-ahead spatial reasoning,'' in \emph{Proceedings of the 63rd Annual Meeting of the Association for Computational Linguistics (Volume 1: Long Papers)}, July 2025, pp. 14\,199--14\,214.

\bibitem{openai2024gpt4ocard}
OpenAI, :, A.~Hurst, \emph{et~al.}, ``Gpt-4o system card,'' 2024.

\bibitem{xiao2024florence}
B.~Xiao, H.~Wu, W.~Xu, \emph{et~al.}, ``Florence-2: Advancing a unified representation for a variety of vision tasks,'' in \emph{Proceedings of the IEEE/CVF Conference on Computer Vision and Pattern Recognition}, 2024, pp. 4818--4829.

\bibitem{vaswani2017attention}
A.~Vaswani, N.~Shazeer, N.~Parmar, \emph{et~al.}, ``Attention is all you need,'' \emph{Advances in neural information processing systems}, vol.~30, 2017.

\bibitem{dit}
W.~Peebles and S.~Xie, ``Scalable diffusion models with transformers,'' in \emph{Proceedings of the IEEE/CVF international conference on computer vision}, 2023, pp. 4195--4205.

\bibitem{yue2024mmmu}
X.~Yue, Y.~Ni, K.~Zhang, \emph{et~al.}, ``Mmmu: A massive multi-discipline multimodal understanding and reasoning benchmark for expert agi,'' in \emph{Proceedings of the IEEE/CVF Conference on Computer Vision and Pattern Recognition}, 2024, pp. 9556--9567.

\bibitem{MMstar}
L.~Chen, J.~Li, X.~Dong, \emph{et~al.}, ``Are we on the right way for evaluating large vision-language models?'' \emph{Advances in Neural Information Processing Systems}, vol.~37, pp. 27\,056--27\,087, 2024.

\bibitem{fu2024mme}
C.~Fu, P.~Chen, Y.~Shen, \emph{et~al.}, ``Mme: A comprehensive evaluation benchmark for multimodal large language models,'' 2024.

\bibitem{guan2024hallusionbench}
T.~Guan, F.~Liu, X.~Wu, \emph{et~al.}, ``Hallusionbench: an advanced diagnostic suite for entangled language hallucination and visual illusion in large vision-language models,'' in \emph{Proceedings of the IEEE/CVF Conference on Computer Vision and Pattern Recognition}, 2024, pp. 14\,375--14\,385.

\bibitem{AI2D}
A.~Kembhavi, M.~Salvato, E.~Kolve, \emph{et~al.}, ``A diagram is worth a dozen images,'' in \emph{European conference on computer vision}.\hskip 1em plus 0.5em minus 0.4em\relax Springer, 2016, pp. 235--251.

\bibitem{team2024realworldqa}
{R. Team}, ``Realworldqa,'' \url{https://x.ai/news/grok-1.5v}, 2024.

\bibitem{duan2024vlmevalkit}
H.~Duan, J.~Yang, Y.~Qiao, \emph{et~al.}, ``Vlmevalkit: An open-source toolkit for evaluating large multi-modality models,'' in \emph{Proceedings of the 32nd ACM international conference on multimedia}, 2024, pp. 11\,198--11\,201.

\end{thebibliography}

\end{document}